\documentclass[sigconf,nonacm,authorversion]{acmart}

\AtBeginDocument{%
  \providecommand\BibTeX{{%
    \normalfont B\kern-0.5em{\scshape i\kern-0.25em b}\kern-0.8em\TeX}}}

\acmSubmissionID{654}

\usepackage{graphicx}
\usepackage{comment}
\usepackage{amsmath} %
\usepackage{color}
\usepackage{verbatim}
\usepackage{wrapfig}
\usepackage[export]{adjustbox}
\usepackage{balance}

\DeclareMathOperator*{\argmax}{arg\,max}

\DeclareMathOperator{\sign}{sign}

\begin{document}
\fancyhead{}

\title{Panoptic Image Annotation with a Collaborative Assistant}

\author{Jasper R. R. Uijlings}
\authornote{Both authors contributed equally to this research}
\email{jrru@google.com}
\affiliation{%
  \institution{Google Research}
}

\author{Mykhaylo Andriluka}
\authornotemark[1]
\email{mykhayloa@google.com}
\affiliation{%
  \institution{Google Research}
}

\author{Vittorio Ferrari}
\email{vittoferrari@google.com}
\affiliation{%
  \institution{Google Research}
}

\begin{abstract}

This paper aims to reduce the time to annotate images for panoptic segmentation, which requires annotating segmentation masks and class labels for all object instances and stuff regions. 
We formulate our approach as a collaborative process between an annotator and an automated assistant who take turns to jointly annotate an image using a predefined pool of segments.
Actions performed by the annotator serve as a strong contextual signal.
The assistant intelligently reacts to this signal by annotating other parts of the image on its own,
which reduces the amount of work required by the annotator.
We perform thorough experiments on the COCO panoptic dataset, both in simulation and with human annotators. These demonstrate that our approach is significantly faster than the recent machine-assisted interface of~\cite{andriluka18acmmm}, and $2.4\times$ to $5\times$ faster than manual polygon drawing.
Finally, we show on ADE20k~\cite{zhou19ijcv} that our method can be used to efficiently annotate new datasets, bootstrapping from a very small amount of annotated data.

\end{abstract}

\bgroup
\tabcolsep 2.0pt
\begin{teaserfigure}
  \hspace{-.7cm}
    \begin{tabular}{cccc}
      \includegraphics[width=0.25\textwidth]{figures/images/000093_110721_image.png}&
      \includegraphics[width=0.25\textwidth]{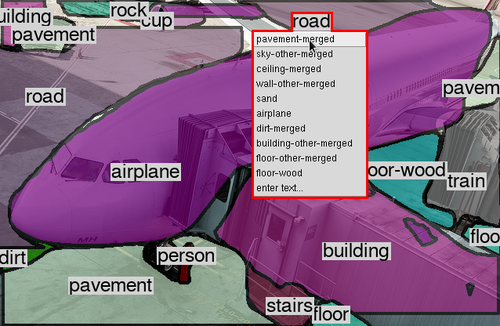}&
      \includegraphics[width=0.25\textwidth]{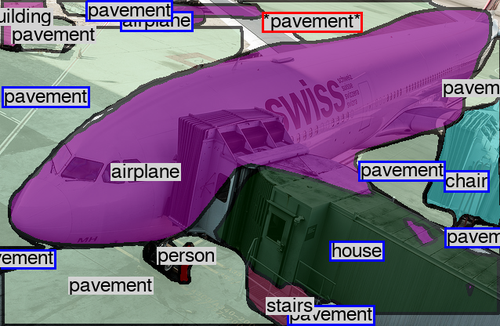}&
      \includegraphics[width=0.25\textwidth]{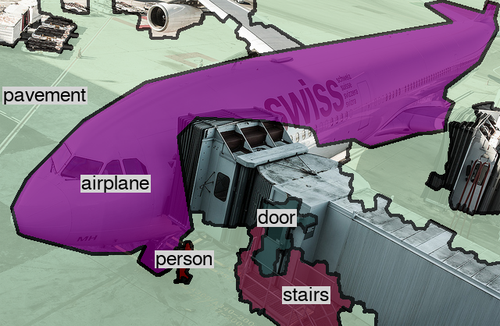}\\
      (a) input image  & (b) the annotator changes &
      (c) the assistant reacts & (d) ground-truth\\
      & \emph{road} to \emph{pavement} & by changing labels & \vspace{0.2cm}\\
      \includegraphics[width=0.25\textwidth]{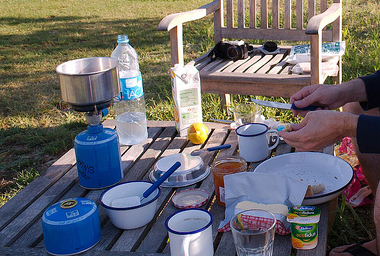}&
      \includegraphics[width=0.25\textwidth]{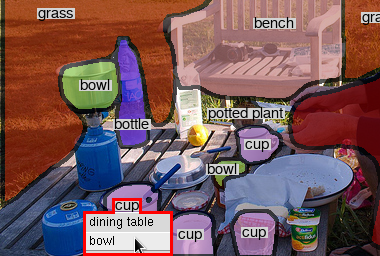}&
      \includegraphics[width=0.25\textwidth]{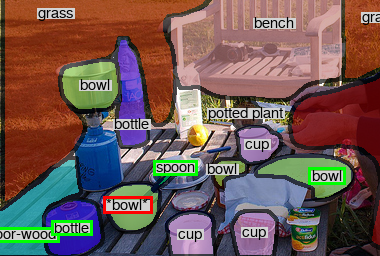}&
      \includegraphics[width=0.25\textwidth]{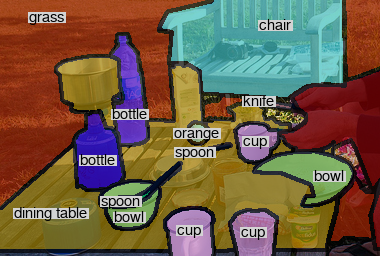}\\
      (e) input image & (f) the annotator changes & (g) the assistant reacts & (h) ground-truth\\
      & \emph{cup} to \emph{bowl} & by adding segments & \\
    \end{tabular}
    \caption{Example of the collaborative annotation process. Given an input image, the human annotator and the automated assistant
      carry out actions in turn. Based on the current annotation, the annotator performs one action (highlighted in red). Then the assistant reacts with by changing labels (in blue) and adding segments (in green). For comparison we show the ground-truth panoptic segmentation.}
    \label{fig:actionexample}
    \vspace{1cm}
\end{teaserfigure}
\egroup

\newcommand{\mypartop}[1]{\vspace{0mm}{\noindent\textbf{#1}}}
\newcommand{\mypar}[1]{\vspace{0mm}{\noindent\textbf{#1}}}

\newif\ifdraft
\drafttrue
\ifdraft
  \newcommand{\misha}[1]{{\color{blue}{M: #1}}}
  \newcommand{\jasper}[1]{{\color{magenta}{J: #1}}}  
  \newcommand{\vitto}[1]{{\color{blue}{V: #1}}}
  \newcommand{\added}[1]{{\color{red}{#1}}}
\else
  \newcommand{\misha}[1]{}
  \newcommand{\jasper}[1]{}
  \newcommand{\vitto}[1]{}
  \newcommand{\added}[1]{}
\fi

\newcommand{\tocheck}[1]{{\color{red}{#1}}}
\newcommand{\changed}[1]{{\color{red}{#1}}}
\newcommand{\agent}{{annotator assistant agent}}
\newcommand{\fixed}{{fixed}}
\newcommand{\fix}{{fix}}
\newcommand{\FA}{Fluid Annotation~}
\newcommand{\COCOPanoptic}{COCO-Panoptic-Challenge~}
\newcommand{\SplitLargeNoSpace}{COCO-58k}
\newcommand{\SplitLarge}{COCO-58k~}
\newcommand{\SplitSmall}{COCO-20k~}
\newcommand{\cls}[1]{\emph{#1}}

\newcommand{\myparbold}[1]{\vspace{-0mm}\paragraph{\textbf{#1}}}

\newcommand{\Eprop}{E_{\mbox{\scriptsize{p}}}}
\newcommand{\Efixavg}{\bar{E}_{\mbox{\scriptsize{fix}}}}
\newcommand{\Efix}{E_{\mbox{\scriptsize{fix}}}}
\newcommand{\Xfix}{X_{\mbox{\scriptsize{fix}}}}
\newcommand{\Kfix}{K_{\mbox{\scriptsize{fix}}}}
\newcommand{\Xprop}{X_{\mbox{\scriptsize{p}}}}

\newcommand{\xfix}{x_{\mbox{\scriptsize{fix}}}}
\newcommand{\xprop}{x_{\mbox{\scriptsize{p}}}}
\newcommand{\changelabel}{\emph{change label~}}
\newcommand{\add}{\emph{add~}}

\begin{abstract}

This paper aims to reduce the time to annotate images for panoptic segmentation, which requires annotating segmentation masks and class labels for all object instances and stuff regions. 
We formulate our approach as a collaborative process between an annotator and an automated assistant who take turns to jointly annotate an image using a predefined pool of segments.
Actions performed by the annotator serve as a strong contextual signal.
The assistant intelligently reacts to this signal by annotating other parts of the image on its own,
which reduces the amount of work required by the annotator.
We perform thorough experiments on the COCO panoptic dataset, both in simulation and with human annotators. These demonstrate that our approach is significantly faster than the recent machine-assisted interface of~\cite{andriluka18acmmm}, and $2.4\times$ to $5\times$ faster than manual polygon drawing.
Finally, we show on ADE20k~\cite{zhou19ijcv} that our method can be used to efficiently annotate new datasets, bootstrapping from a very small amount of annotated data.

\end{abstract}

\maketitle

\section{Introduction}

This paper aims to reduce the time it takes to annotate images for the panoptic segmentation task. This requires annotating segmentation masks and class labels for all object instances and stuff regions.
Such annotations are expensive: it took 19 minutes for a single image for COCO~\cite{caesar18cvpr,lin14eccv} and 1.5 hours for Cityscapes~\cite{cordts16cvpr}.
In this paper we propose to reduce annotation time by learning to predict how the image should be annotated.

To this end we formulate our approach as a collaborative process between a human annotator and an automated assistant agent who take turns to jointly annotate an image.
To see this process in action, Fig.~\ref{fig:actionexample}a shows an example image and Fig.~\ref{fig:actionexample}b the current annotation, which is partially machine generated.
At this point, the annotator converts one of the \cls{road} segments into \cls{pavement}. The assistant reacts by changing the labels of similar looking segments elsewhere in the image to \emph{pavement} as well.
Fig.~\ref{fig:actionexample}e and~\ref{fig:actionexample}f show another example image and current annotation. After the annotator corrects the \cls{cup} to be a \cls{bowl}, the assistant automatically adds another \cls{bowl}, as well as a \cls{spoon}, a \cls{bottle}, and a \cls{wooden floor}. This significantly helps the annotator.

We build on~\cite{andriluka18acmmm}, which proposed to annotate an image by composing segments out of a pre-defined pool, using an interface in which annotators can repeatedly perform one of the following actions: add a segment from the pool, change the label of a segment, and remove a segment.
In this paper we introduce an automatic assistant which helps the annotator by executing some of these actions on its own. Crucially, whenever the annotator performs an action, this directly provides a ground-truth annotation for a segment. This ground-truth serves as a strong form of context, stronger than the use of {\em predicted context} in previous works~\cite{heitz08eccv,hu19nn,lin18eccv,modolo15bmvc,murphy03nips,rabinovich07iccv,tighe11iccv}.
We propose an assistant that reacts  intelligently to annotator actions by capitalizing on these strong contextual cues to automatically improve the annotation of other parts of the image.

To summarize, we introduce a framework in which an assistant and an annotator collaboratively annotate an image. The assistant intelligently reacts to annotator input based on context by annotating other parts of the image on its own.
We experimentally demonstrate:
(1) in simulations on the COCO panoptic dataset~\cite{caesar18cvpr,kirillov19cvpr,lin14eccv} our method is $17\%-26\%$ faster than~\cite{andriluka18acmmm};
(2) studies with human annotators on this dataset confirm this and enable to compare to manual polygon drawing,
revealing that our method is $5 \times$ faster at a small loss of quality.
Furthermore, after adding a manual refinement stage, our overall pipeline is $2.4 \times$ faster without any compromise on quality.
(3) a cross-dataset experiment on ADE20k~\cite{zhou19ijcv} shows that our method can be used to quickly annotate new datasets, bootstrapping from a very small amount of labeled data.

\section{Related Work}

\mypar{Interactive Segmentation.}
Many works address interactive object segmentation. Most classical approaches ~\cite{bai09ijcv,batra11ijcv,boykov01iccv,rother04siggraph,criminisi10tog,cheng15cgf,gulshan10cvpr,nagaraja15iccv} cast the problem as an energy minimization function defined on a graph which spans over pixels. Many recent methods adapt Fully Convolutional neural networks (FCNs, e.g. ~\cite{chen18pami,long15cvpr}) for single object segmentation by using user scribbles or clicks as additional input signal~\cite{benenson19cvpr,hu19nn,le18eccv,li18cvpr,liew17iccv,mahadevan18bmvc,maninis18cvpr,xu16cvpr}.
In~\cite{chen18cvpr} they use an FCN to produce pixel-wise embedding space. They combine this with user clicks and a nearest neighbour classifier to segment objects within video.
Polygon-RNN~\cite{acuna18cvpr,castrejon17cvpr} is a recurrent neural net which predicts polygon vertices which an annotator can adjust.
In~\cite{le18eccv} they predict object boundaries using an FCN which accepts boundary clicks, which they turn into an instance using a geodesic path solver~\cite{cohen96cvpr}.

While the vast majority of works focuses on individual objects, a few recent works have started to tackle the full image annotation problem~\cite{andriluka18acmmm,agustsson19cvpr}.
In~\cite{agustsson19cvpr} they do class-agnostic segmentation of all objects in an image starting from extreme points~\cite{papadopoulos17iccv} followed by corrective scribbles.
The closest related work to ours is Fluid Annotation~\cite{andriluka18acmmm}. Instead of segmenting one object at a time, they propose an interface to quickly annotate a complete image by composing segments out of a pre-defined pool. The interface is designed to make it easy for the annotator to perform the action she chooses. In this work we build on top of~\cite{andriluka18acmmm} and go beyond by introducing an automated assistant that performs some annotation actions on its own.

\mypar{Assign annotation tasks.}
A few works propose to intelligently choose what annotation task to send to the annotator~\cite{konyushkova18cvpr,vijayanarasimhan09cvpr,russakovsky15cvpr}. 
In~\cite{konyushkova18cvpr} they focus on creating a bounding box for each image-label pair. They have an agent decide whether to ask the annotator to draw a bounding box~\cite{papadopoulos17iccv,su12aaai} or verify whether a machine-generated bounding box is good enough~\cite{papadopoulos16cvpr}. In~\cite{russakovsky15cvpr} the machine dispatches annotation tasks to optimize the trade-off between the annotation budget and the final labeling quality. Tasks include providing image labels, providing a label for a certain box, verify a box, draw boxes around other instances of the same class, etc. In~\cite{vijayanarasimhan09cvpr} they estimate the informativeness and cost of having an image label, a box, or a full segmentation of an image, which they use to dispatch these tasks in an active learning framework.
In these works the machine decides which tasks to send to the annotators while the tasks themselves are not interactive. In our case there is a single interface through which both the machine and the annotator interactively take turns.

\mypar{Other works on interactive annotation.}
In~\cite{rupprecht18cvpr} they created a semantic segmentation network which can consume natural language input. This enables a user to correct a machine-predicted segmentation by giving instructions in English. In~\cite{branson14cvpr} they propose a framework to convert bounding boxes, instance masks, and part annotations into each other using human verification and human corrections. Other works address fine-grained classification, where annotators provide feedback or correct the classifiers based on attributes~\cite{branson10eccv,parkash12eccv,biswas13cvpr,wah14cvpr}.

\section{Method}

Given an input image we want to produce a dense labelling of every pixel with a semantic label and
object identity. This includes ``thing'' classes corresponding to various
countable objects, and ``stuff'' classes corresponding to uncountable classes which typically occupy
background areas.
Example annotations are shown in Fig~\ref{fig:actionexample}d and~\ref{fig:actionexample}h. 

We start from the recent \FA interface \cite{andriluka18acmmm} that allows to quickly annotate an image by composing segments out of a pre-defined pool (Sec.~\ref{sec:FA}). In this paper we turn this into a collaborative environment (Sec.~\ref{sec:collaborative}) and introduce an automated assistant which helps the annotator complete its task (Sec.~\ref{sec:assistant}).
Crucially, every action of the annotator provides strong contextual cues which the assistant uses to predict how the image should be annotated. Then the assistant \emph{carries out some actions on its own}.

\subsection{Fluid Annotation~\cite{andriluka18acmmm}}\label{sec:FA}

\FA\cite{andriluka18acmmm} starts by generating a \emph{proposal set} of segments with accompanying class labels using Mask-RCNN~\cite{he17iccv}. They modified Mask R-CNN to generate about 1000 segments per image (more than usual)
and to produce segments also on stuff classes (not only on things).
The image is annotated by selecting an ordered subset of these proposals and by optionally correcting their labels. The proposals are ordered to ensure that a single pixel is assigned to only one segment, following the definition of panoptic segmentation~\cite{kirillov19cvpr}.

\begin{figure}[t]
\vspace{-.4cm}
    \centering
      \includegraphics[width=.89\linewidth]{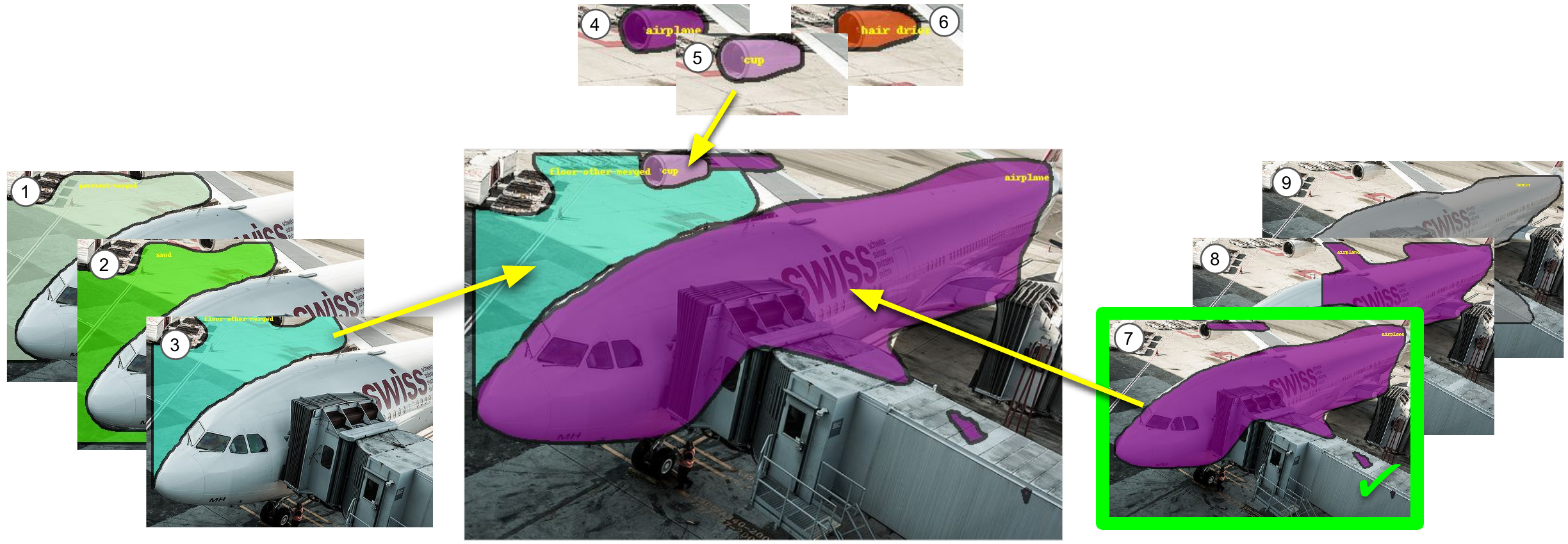}\\
      \vspace{-.0cm}
      \caption{The \emph{proposal set}, \emph{active set} and \emph{fixed set}. The \emph{proposal set} is the complete segment pool created by Mask-RCNN (few are shown for clarity). The \emph{active set} defines the current annotation (segments 3, 5, 7). The \emph{fixed set} contains segments which were modified/added by the annotator and are considered to be ground-truth (segment 7). %
      }
    \label{fig:illustration_proposal_set}
\end{figure}

From the proposal set, the work of ~\cite{andriluka18acmmm} first creates an initial annotation for the whole image using an iterative greedy algorithm~\cite{andriluka18acmmm,kirillov19cvpr}. Starting from the empty image, first the highest scored segment is selected. Then the next-highest scored segment is placed behind all other segments, if enough of its surface is visible.
This algorithm creates an initial ordered \emph{active set}. This active set is a subset of the proposal set, and defines the current annotation (Fig.~\ref{fig:illustration_proposal_set}).
The annotator can modify the active set by performing four kinds of actions: \emph{add a segment} from the proposal set into the active set,
\emph{remove a segment},
\emph{change label} of an active segment,
and \emph{change depth order} of an active segment.

The \FA interface facilitates the actions made by the annotator: when adding a segment, the mouse position is used to make a small, ordered selection of segments for the annotator to quickly scroll through. When changing a label, the interfaces makes a shortlist of likely labels for that segment. Hence in~\cite{andriluka18acmmm} the system facilitates the annotator to perform the one action she chose.

\subsection{Collaborative fluid annotation}
\label{sec:collaborative}

In this paper we want to extend the influence of each annotator action beyond the one segment that was targeted. We achieve this by introducing an {\em automatic assistant} to the \FA system.
We model annotation as a collaborative environment in which the annotator and the assistant alternate in taking turns, where both use the same set of actions.
Conceptually, the annotator has perfect
knowledge about the visual world and about the annotation it aims to achieve. However, it has only a
partial view of the proposal set, and exploring this set is a costly process.
Conversely, the assistant can access all the proposal segments instantly, but has limited capability to judge which proposals belong to the final segmentation the human annotator would like to achieve.
The aim of the annotator is to produce a high-quality panoptic annotation. The goal of the assistant is to reduce the overall annotation effort.

Crucially, the annotator conveys his knowledge about the world through every action she takes, essentially creating ground-truth as she goes. This freshly created knowledge provides strong contextual cues which the assistant uses to predict how the rest of the image should be annotated. This enables the assistant to react to the annotator and carry out some actions on its own.

\mypar{Fixed Set.}
To establish communication between the annotator and the assistant we introduce a \emph{fixed set}.
It contains all segments which have been approved by the annotator and are considered to be ground-truth. It is a subset of the active set, which in turn is a subset of the proposal set (Fig.~\ref{fig:illustration_proposal_set}).
Since the fixed set can be considered as ground-truth, we do not allow the assistant to make any changes to it.

To create the fixed set, we make four natural assumptions on the behavior of the annotator:
(1) whenever the annotator changes the label of a segment, they set it to the correct label;
(2) the annotator only changes the label of a geometrically correct segment (i.e. matching well a real object or background region);
(3) the annotator only adds segments that are geometrically correct;
(4) whenever an annotator adds a segment which has an incorrect label, they will correct it with their next action.
Using these assumptions, we can now create the fixed set automatically without additional costs.
By combining (1) and (2), if the annotator changes a label of a segment, we immediately put it into the fixed set.
By combining (3) and (4), if an annotator adds a segment, we put it into the fixed set after their next action.
Using the logs of our human experiments (Sec.~\ref{sec:human}), we found that these assumptions hold in the vast majority of the cases (90\%-96\%).
Moreover, these assumptions only affect the fixed set and hence the behavior of the assistants, {\em not} the actual behavior of the annotator.

\label{sec:assistant}

\mypar{Workflow.}
The assistant and the annotator take turns to collaboratively annotate an image. In particular, the assistant performs as many actions as it wants to, until it decides stop. Then the annotator performs one action.
If this action alters the fixed set, the assistant has more information which it can use during its next turn.

We introduce a collaborative assistant in Sec.~\ref{sec:context_assistant}. This assistant reacts to context provided by the annotator and can perform the \emph{change label} and \emph{add} actions.
We also introduce an initialization agent in Sec.~\ref{sec:initialization_assistant}. It acts before the annotator and replaces the greedy initialization stage of \FA (Sec.~\ref{sec:FA}).

\subsection{Collaborative Assistant (CA)}
\label{sec:context_assistant}
Our collaborative assistant automatically performs actions to accelerate the annotation process. 
To identify useful actions, it relies on a \textit{context model} that
captures dependencies between fixed segments and the segments in the proposal set.
Here we describe the action generation process of the assistant, and then present details of the context model.

\mypar{Action generation.}
Denote the fixed set as $\Xfix$. For the \emph{change label} action, our context model predicts $p(c_p^k|\Xfix)$ for each segment $k$. The collaborative assistant generates a \changelabel action when the highest scored label of the context model is different from the current label. The action updates the label to $m$, using $\argmax_{m}p(c_p^k = m|\Xfix)$.

\bgroup
\tabcolsep 15pt
\begin{figure*}[t]
\vspace{-0.0cm}
    \centering
    \begin{tabular}{ccc}
      \includegraphics[width=0.25\textwidth]{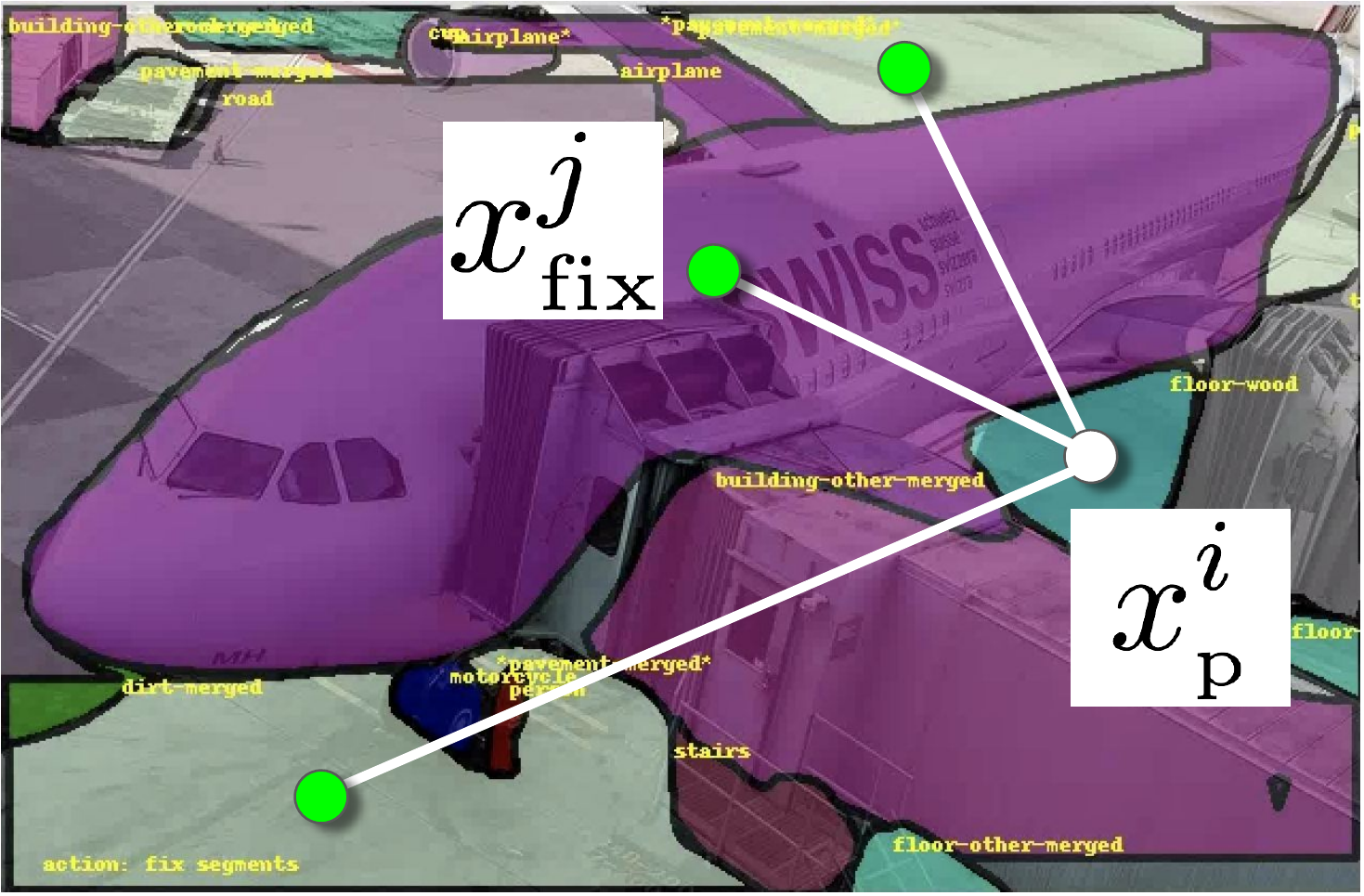}&
      \includegraphics[width=0.22\textwidth]{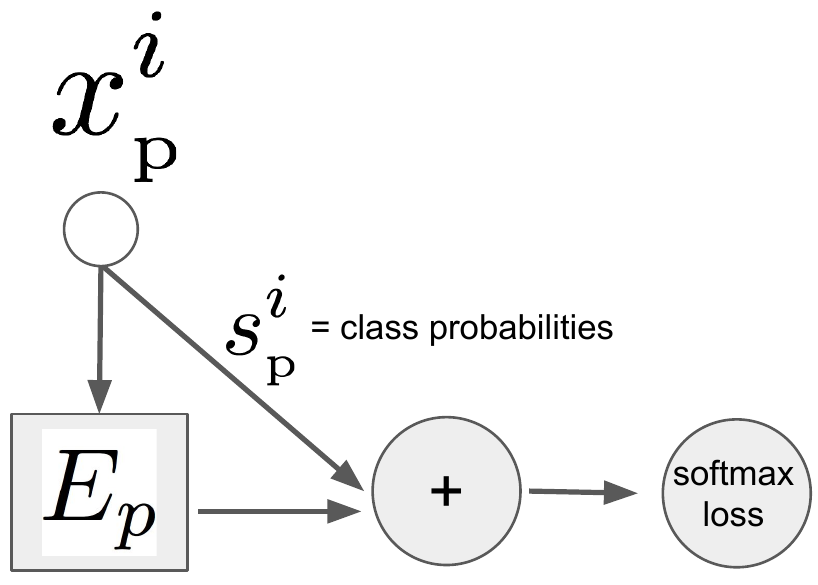}&
      \includegraphics[width=0.38\textwidth]{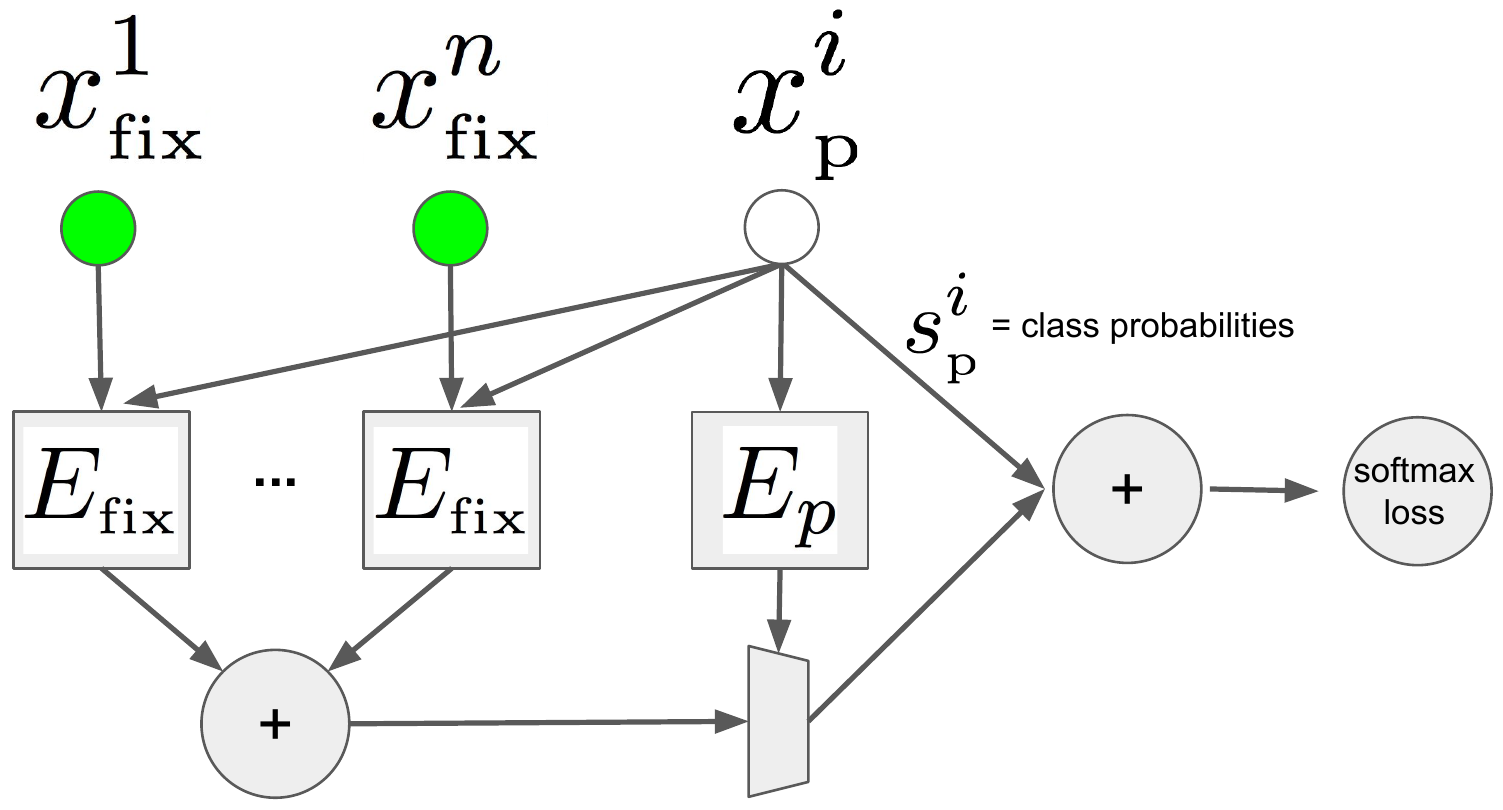}\\
      (a) & (b) & (c) \\\vspace{-.8cm}
    \end{tabular}
    \caption{(a) Example image with three fixed segments (green dots), and a
      proposal segment (white dot). (b) Diagram of a simple model with skip connections that does not
      have access to fixed segments. (c) Diagram of our context model.}
    \label{fig:contextmodel}
\end{figure*}
\egroup

\begin{figure}[t]
    \centering
    \includegraphics[width=0.9\linewidth]{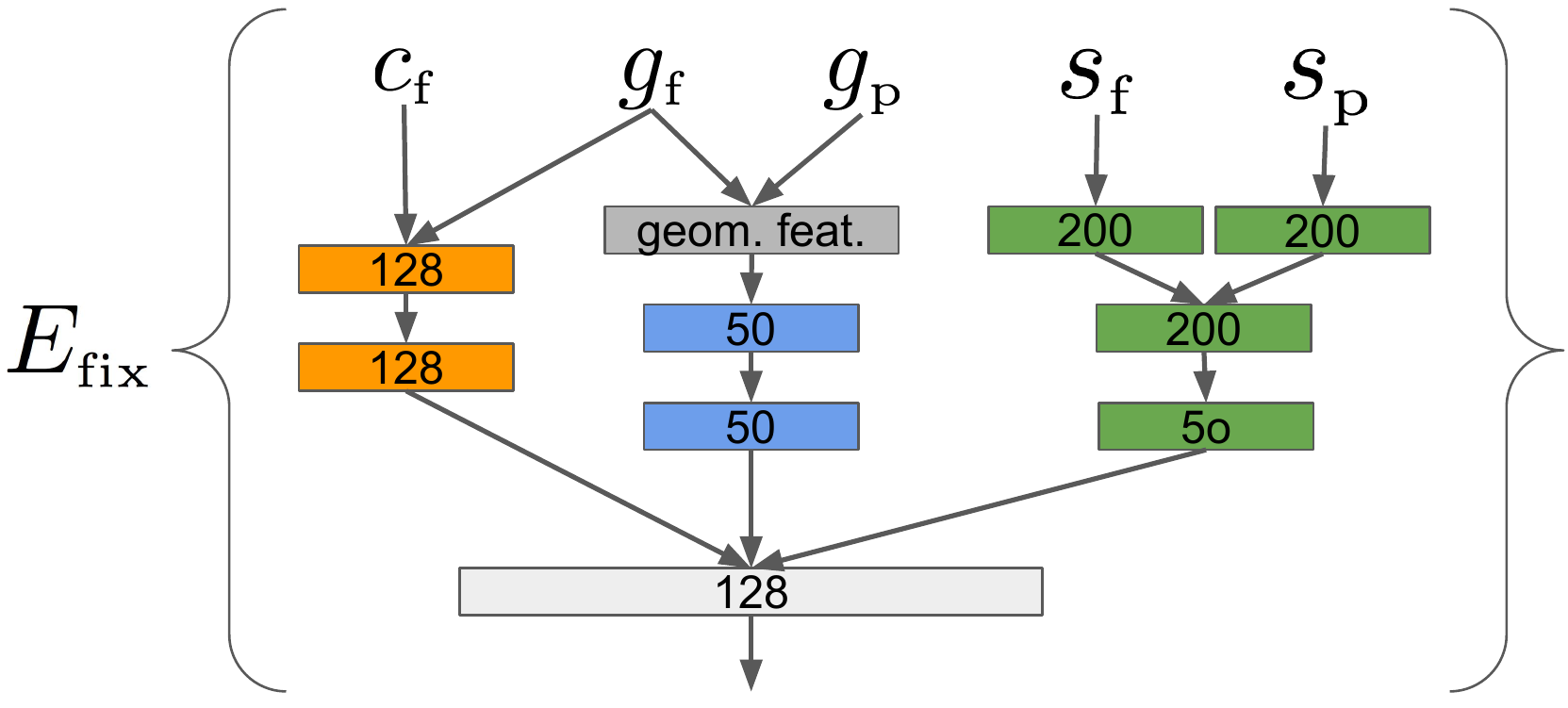}
    \vspace{-.4cm}
    \caption{Structure of the $\Efix$ subnetwork: 
    each block is a fully connected layer with ReLu activation, where the number specifies the number of channels. Features specifying 
  human-provided inputs, geometry, and appearance are first separately processed (orange, blue, and green respectively) and fused in the final layer (light gray). All feature fusion is done through concatenation.}
    \label{fig:efix_subset}
\end{figure}

Similarly, for the \emph{add} action our model predicts $p(d_p^k|\Xfix)$.
The \add action is generated whenever $p(d_p^k|\Xfix)$ is 
greater than threshold $\tau$ and segment $k$ is not yet in the active set. We set $\tau=0.9$ as estimated on a validation set (Sec.~\ref{sec:results}). 

\mypar{Context model.}
For both actions we use a context model with the same architecture. 
Let the set of fixed segments be denoted by
$\Xfix = \{\xfix^k|k=1, \ldots, K\}$ and the set of proposal segments by
$\Xprop = \{\xprop^i|i=1, \ldots, N\}$.  

In our model the features of the fixed and proposal segments are given by
$\xfix^k=[c_f^k, g_f^k, s_f^k]$ and $\xprop^i=[g_p^i, s_p^i]$ respectively. Each of the variables $g_f^k$
and $g_p^i$ corresponds to a 4D vector encoding the center of the segment and the
width and height of its bounding box. This captures spatial relationships between the proposals.
The variables $s_f^k$ and $s_p^i$ each correspond to a vector of
class scores (i.e. logits) assigned to the segment by Mask-RCNN. These logits capture semantic information and, since they are linear projection of the final feature vectors, they also contain information about the segments' appearance.
Finally, the variable $c_f^k$ denotes a one-hot
encoding of the segment class that has been assigned to the fixed segment by the human annotator.
Note that the $c_f^k$ component of the fixed segment representation $\xfix^k$  provides an additional and potentially
strong cue for resolving ambiguities in the label of the proposal segments.
This is a new piece of information not accessible to Mask-RCNN as usually deployed (i.e. without a collaboration with a human).

The structure of our \emph{change label} context model is shown in Fig.~\ref{fig:contextmodel}c.
Our starting point is a simple update model shown in Fig.~\ref{fig:contextmodel}b. This model takes the segment features $\xprop^i$ as input and
computes incremental updates to the class probabilities using a fully connected neural network $\Eprop$, similarly to a single layer in ResNet~\cite{he16cvpr}.
We extend this simple model with a graph-convolutional~\cite{kipf17iclr} sub-network that computes a fixed-size vector summarizing the
relationship between the proposal and the unordered, arbitrarily sized, fixed set:
\begin{equation}
\Efixavg = \frac{1}{K}\sum_{k}\Efix(\xprop^i, \xfix^k).
\label{eq:avgpool}
\end{equation}
The output of $\Efixavg$ is then combined with the output of $\Eprop$ and is passed though a single
fully connected layer to compute a difference vector with respect to the original class
probabilities of a proposal segment $x_p^i$. 

Our context model for the \emph{add} action is almost identical in structure. But instead of updating all class scores $s_p^i$ it only updates the highest class score while the softmax loss is replaced by a binary cross-entropy loss (present/absent).

In Fig.~\ref{fig:contextmodel} (d) we show the structure of the component $\Efix$ that encodes
relationship between a fixed segment and a proposal segment. We follow the late fusion strategy and first
apply a series of transformations to each type of input features before finally combining them
together. Prior to feeding geometric features $g_f$ and $g_p$ into the network, we transform them
into a 10-dimensional vector of relative features defined similarly to \cite{hu18cvpr}. Let us
denote the offset vector between the locations of the fixed and the proposal segment as
$\Delta_{fp}$, and after normalization by the width and height of the fixed segment as
$\hat{\Delta}_{fp}$. The vector of relative geometric features is then obtained by concatenating
$\Delta_{fp}$, $\log(|\hat{\Delta}_{fp}|)$, $\log(w_p/w_f)$, $\log(h_p/h_f)$, $\sign(\Delta_{fp})\log(|\Delta_{fp}|)$,
and $\sign(\Delta_{fp})\log(|\hat{\Delta}_{fp}|)$.

\mypar{}

\mypar{Local score pooling.}
The model described above can operate with any type of segment scores
$s_p^i$.
We found that instead of directly using the scores provided by Mask-RCNN we achieve higher
accuracy by performing a form of max-pooling over the scores of nearby segments.
This corresponds to defining a new score vector $\hat{s}^i_p$ as
\begin{equation}
\hat{s}^i_{p,c} = \max_{j \in N(i,c)}s^j_{p,c}
\label{eq:local_pooling}
\end{equation}
where $c$ is a class label, and $N(i,c)$ is the set of segments with label $c$ that have intersection-over-union $<0.5$ with segment $i$.

\mypar{Training the context model.}
The ideal training data for our context model would be composed of training
examples derived from the actions performed by real humans during image annotation. Since collection of
such training data is impractical we use data collected in simulation.
To that end we simulate the fluid annotation process for each of the images in the training set using the simulator described in \cite{andriluka18acmmm} and store all segments and their labels contained in the final annotation.
To construct training examples of the fixed set for our context model we then randomly sample correct segments out of the final annotation.
We found such random sampling to be necessary to make the model robust with respect to the size of the fixed set encountered by the agent when deployed at annotation time. We train the context model using the Adam optimizer \cite{kingma15iclr}.

\mypar{Discussion.}  
A number of previous works attempts to exploit context based purely on image signals~\cite{heitz08eccv,hu19nn,lin18eccv,murphy03nips,rabinovich07iccv,tighe11iccv}. Instead, our contextual cues are based on \emph{human input},
which makes them much stronger.
In terms of technical realization, our context model is related to models used for visual question
answering \cite{santoro17nips} and for modeling relationships between scene objects
\cite{hu18cvpr}.

\subsection{Initialization Assistant (IA)}
\label{sec:initialization_assistant}

In the original Fluid Annotation, the initialization was done greedily through non-maximum suppression of Mask-RCNN segments based on their scores~\cite{andriluka18acmmm}.
In this section we use an initialization assistant to do this instead. The initialization produces a panoptic segmentation by composing segments from the proposal set, without any human annotator involved. This method therefore can also be used for classical image segmentation prediction \cite{chen18pami,kirillov19cvpr,long15cvpr,mottaghi14cvpr}.

We represent each segment as a feature vector consisting of:
(a) an 8-bit encoding  of the class label predicted by Mask-RCNN;
(b) the predicted score of the segment;
(c) the percentage of the segment surface that does not overlap with any segments in the current active set.
The assistant uses these features to iteratively perform actions, starting from an empty active set. At each time step, it scores all the segments which could be added and select the one with the highest score. If this score is above a certain threshold, it adds the segment to the active set. Otherwise, it stops.

To train our initialization assistant, we collect examples by simulating the full fluid annotation process for each image (as in Sec.~\ref{sec:context_assistant}). This results in a set of positive \emph{add} actions with their features. We obtain negative \emph{add} actions with a form of hard negative mining. In particular, at several points during the fluid annotation simulation we ask the initialization assistant to make predictions. All predicted \emph{add} actions which are inconsistent with the original ground truth form the negative example set. As model we use a simple 4-layer fully connected neural net. We use a quadratic hinge loss and train using the Adam optimizer~\cite{kingma15iclr}.

Our initialization assistant is conceptually related to the search-based structured prediction algorithm~\cite{daume09ml}, and to other approaches that iteratively generate structured outputs~\cite{gkioxari16eccv,banica15cvpr}.

\section{Results}
\label{sec:results}

In this section we first evaluate the context model (Sec.~\ref{sec:eval_context}) and the full collaborative annotation process in simulation (Sec.~\ref{sec:eval_full}).
Then we report a study with human annotators (Sec.~\ref{sec:human})
and report a cross-dataset experiment (Sec.~\ref{sec:cross_data}).

\mypar{Dataset.}
In most experiments we use the COCO panoptic dataset\cite{cocopanoptic}, which combines the original COCO dataset~\cite{lin14eccv} with COCO-stuff~\cite{caesar18cvpr}, while merging some stuff classes based on~\cite{kirillov19cvpr}.
It contains 118K training and 5K validation images densely labeled with 80 thing and 53 stuff classes.
We divide the training set into two equal sized splits, one to train Mask-RCNN and another for our assistants (including the context models). We use the validation set for evaluation. 

In Sec.~\ref{sec:cross_data} we use ADE20k to demonstrate that our method can be used to annotate a new dataset (detailed protocol in Sec.~\ref{sec:cross_data}).

\subsection{Context model}
\label{sec:eval_context}

\begin{figure}[t!]
  \vspace{-.3cm}
    \includegraphics[width=\linewidth]{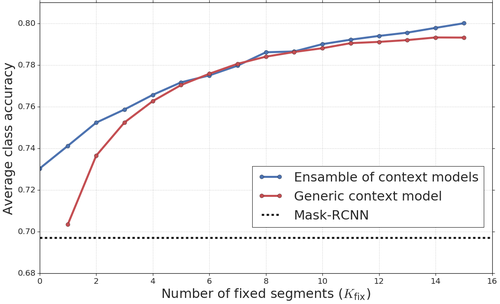}
    \caption{Class label accuracy of our context model (red, blue) w.r.t. the number of fixed segments given to it. The Mask-RCNN baseline (dashed black) represents the starting point without context model. Note that for $\Kfix = 0$ the accuracy of the ensemble model is higher than the baseline due to local score pooling. (Eq.~\eqref{eq:local_pooling}).}
    \label{fig:change_label_results60k}
  \vspace{-.3cm}
\end{figure}

In this section we evaluate the performance of our context model in isolation. We measure how well it can correct labels predicted by Mask-RCNN, given a set of fixed segments $\Xfix$.

\begin{figure*}[t!]
\vspace{-0.0cm}
  \begin{minipage}{0.6\textwidth}
\includegraphics[width=\linewidth]{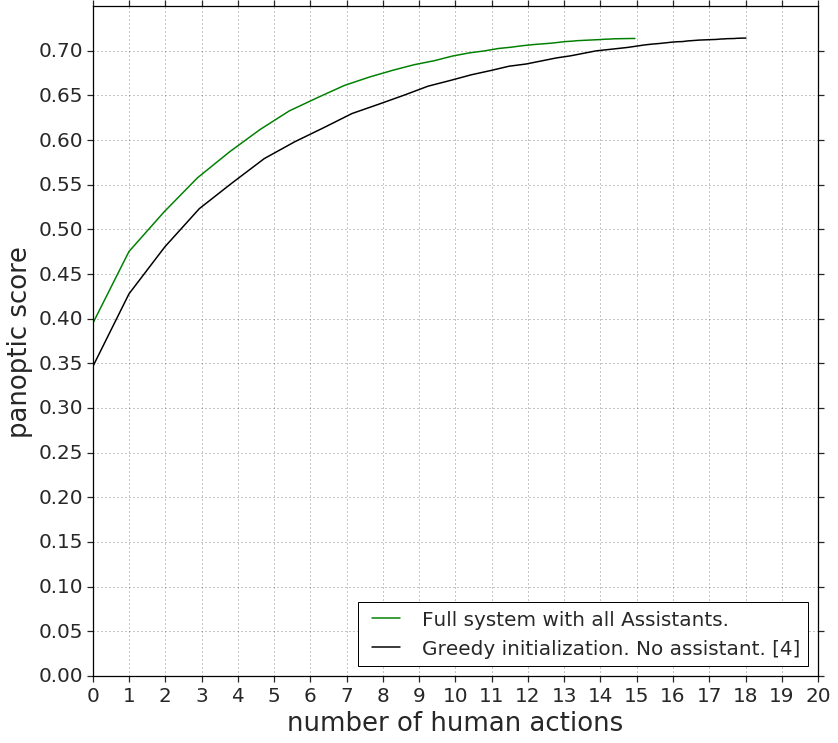}
\vspace{-.0cm}
\caption{Trade-off between effort and quality for our annotation system on the COCO panoptic dataset~\cite{caesar18cvpr,kirillov19cvpr,lin14eccv} (in simulation). The black line shows the baseline system~\cite{andriluka18acmmm}. The green line is our full system.} %
\label{fig:full_system_results2}
\end{minipage}
  \hspace{0.7cm}
  \begin{minipage}{0.315\textwidth}
  \centering
  \vspace{-.0cm}
  \includegraphics[width=0.95\linewidth]{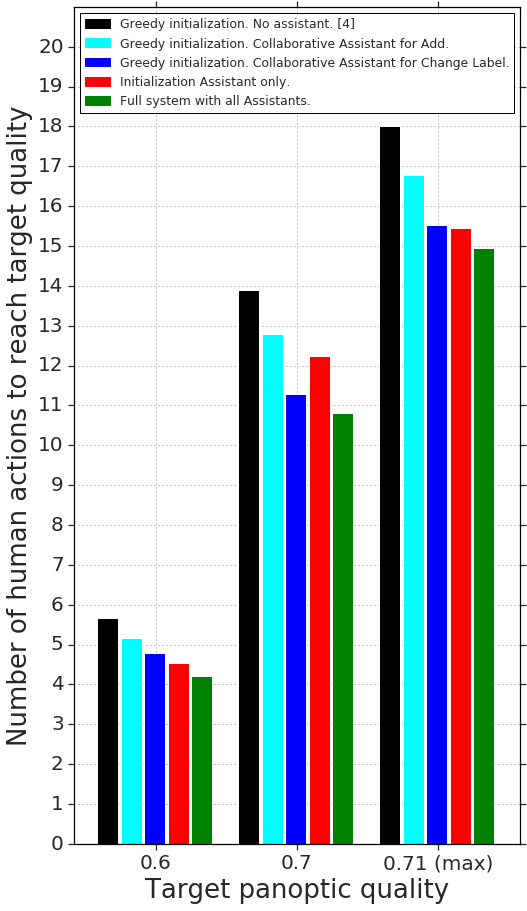}
  \vspace{-.2cm}
  \caption{Ablation study visualizing the number of actions required to reach a certain target quality.} 
  \label{fig:ablation}
\end{minipage}
\end{figure*}

\mypar{Protocol.}
Given an image and the ground-truth segmentation, we first select the set of Mask-RCNN proposals $E$ which best match the ground-truth segments geometrically. We do this by greedily selecting proposal segments that have maximal IoU with a ground-truth segment.
We then measure the initial label accuracy provided by Mask R-CNN on these selected proposals $E$, and compare it to the accuracy after re-labeling them with the context model. We do this on 4500 COCO validation images (out of 5000) and measure accuracy averaged over classes.

When applying our context model, for each segment $s \in E$ we randomly select a fixed set of other segments in $\Xfix \in E \setminus s$. We evaluate over a range of different number of fixed segments $|\Xfix|$ and each time measure accuracy over all other segments. For this experiment we train a single \emph{generic context model} which works for any size of the fixed set. Since this may be sub-optimal, we also train a separate context model for each specific fixed set size, and combine them into an \emph{ensemble model}.

\textbf{Results}. As Fig.~\ref{fig:change_label_results60k} shows, our context model helps improve label prediction accuracy, and its effect grows with the number of fixed segments given to it. For example, when given 8 fixed segments the accuracy increases from 69\% for the original Mask-RCNN to 78\% with our context model. This demonstrates that 
the fixed segments provide a strong context signal.
We note that this effect is greater than typically demonstrated in classical context works, which do not benefit from conditioning on human input for parts of image~\cite{heitz08eccv,hu19nn,lin18eccv,modolo15bmvc,murphy03nips,rabinovich07iccv,tighe11iccv}.

We also observe that our ensemble model always performs better or equal to our generic model, where the difference is especially large for $\Kfix \leq 3$. The ensemble model takes into account the size of $\Kfix$ and exploits the fact that, intuitively, the more fixed segments are given, the stronger the context signal and thus the more it can be relied upon to alter the labels of other segments.

\subsection{Collaborative annotation process}
\label{sec:eval_full}

We now evaluate our assistant in the full collaborative annotation environment.
We simulate an annotator that tries to reproduce the original ground-truth of the COCO panoptic challenge~\cite{caesar18cvpr,lin14eccv}.
To keep the experiment fully rigorous, we do not report results on the 4500 validation images used to evaluate the context model, but on the remaining 500 images instead.

\mypar{Results.}
Fig.~\ref{fig:full_system_results2} measures quality (panoptic score~\cite{kirillov19cvpr}) as a function of annotation effort (number of human actions). We compare the original \FA system~\cite{andriluka18acmmm} (i.e. greedy initialization and no assistants) with our system which includes all assistants. As can be seen, our system is significantly better than~\cite{andriluka18acmmm} across the whole trade-off curve between effort and quality.
In particular, to reach 0.60 panoptic quality, our system needs 4.2 actions compared to 5.7 actions of~\cite{andriluka18acmmm}, a speed-up of 26\%. Our system reaches the final quality of 0.71 after 14.9 annotator actions vs. 18.0 required by~\cite{andriluka18acmmm}, a speed-up of 17\%.
From an alternative perspective, if we fix the annotation budget to 6 annotator actions, our system yields 4\% higher panoptic quality.
Finally, we note that our approach is several times more efficient than the traditional annotation strategy of manually drawing polygons~\cite{cordts16cvpr,lin14eccv,russell08ijcv,zhou19ijcv}, as we demonstrate through human studies in Sec.~\ref{sec:human}.

\hspace{-2.0cm}
\begin{figure*}[htpb]
  \vspace{-0.0cm}
  \setlength\tabcolsep{.5pt}
  \begin{tabular}{cccc}
      \includegraphics[width=.25\linewidth]{figures/examples_full_system/128748__im.png} &
      \includegraphics[width=.25\linewidth]{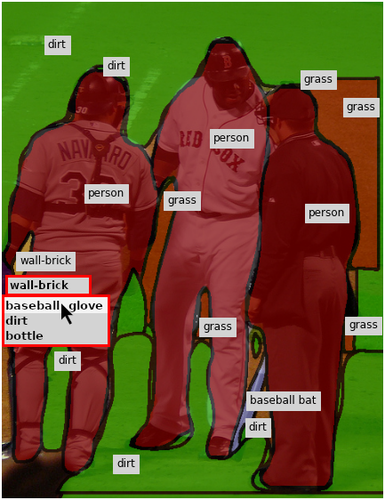} &
      \includegraphics[width=.25\linewidth]{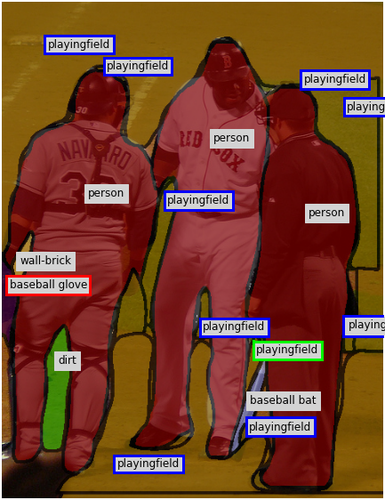} &
      \includegraphics[width=.25\linewidth]{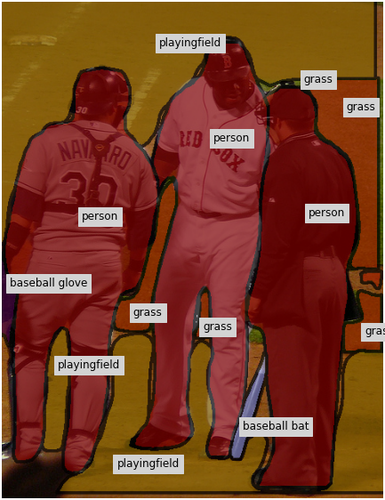} \\
      \scriptsize{(a) original image} & \scriptsize{(b) annotator changes \emph{wall} to \emph{baseball glove}} & \scriptsize{(c) assistant changes other parts of the image} & \scriptsize{(d) final annotation result} \\
      \includegraphics[width=.25\linewidth]{figures/examples_full_system/308793__im.png} &
      \includegraphics[width=.25\linewidth]{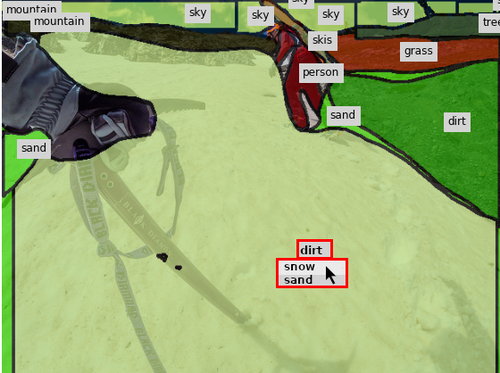} &
      \includegraphics[width=.25\linewidth]{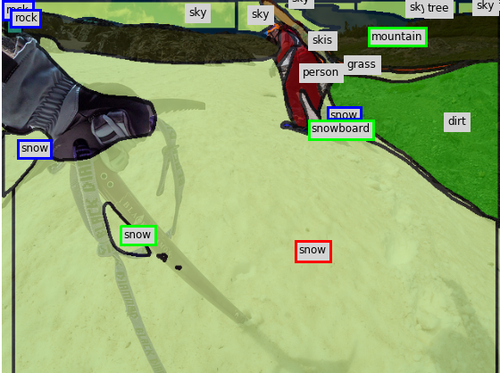} &
      \includegraphics[width=.25\linewidth]{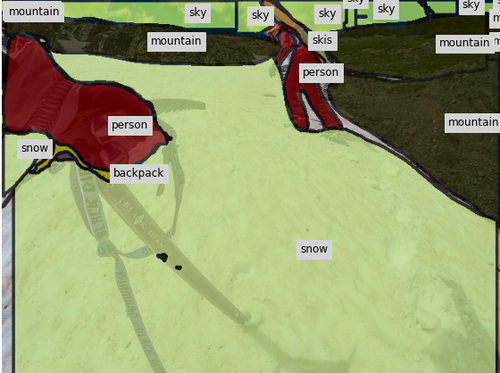} \\
      \scriptsize{(e) original image} & \scriptsize{(f) annotator changes \emph{dirt} to \emph{snow}} & \scriptsize{(g) assistant changes other parts of the image} & \scriptsize{(h) final annotation result} \\
  \end{tabular}
  \caption{Two examples of the annotation process. When the annotator changes a label (highlighted in red), the assistant reacts by changing labels of existing segments (in blue) and adding new segments (in green).}
  \label{fig:examples_full_system_1}
\end{figure*}

\mypar{Ablation.}
To gain insights into the contributions of each assistant, we combine each assistant individually with the baseline~\cite{andriluka18acmmm}. We show in Fig.~\ref{fig:ablation} the number of human actions required to reach a desired target quality (the fewer, the better). Note that all systems eventually reach the same final panoptic quality of 0.71.
We conclude the following:
(1) Each individual assistant brings visible improvements over~\cite{andriluka18acmmm} alone.
(2) The assistant which performs the \emph{change label} action is consistently better than the one performing \emph{add segment}. Intuitively, this makes sense since changing the label of an existing segment is easier than adding an entirely new segment.
(3) While the initialization assistant gives the most improvements after few human actions (see target quality = 0.6), the assistant which performs \emph{change label} gives the most improvements when more human actions are performed (see target quality = 0.7). This makes sense since the initialization assistant does not do anything during interaction, while the \emph{change label} assistant receives an increasing amount of context signal which it can act upon.
(4) Our full system consistently improves over using each assistant individually, demonstrating that our assistants are complementary and all contribute to the final speed improvements.

We can also measure the influence of the assistants in terms of how much they reduce specific types of human actions (comparing the baseline~\cite{andriluka18acmmm} with our full system).
The change label assistant reduces the number of {\em human} change label actions by 29\%.
The initialization assistant and the add segment assistant together reduce the combined number of {\em human} add and remove segment actions by 12\% (they affect both add and remove, since wrongly added segments need to be removed).

\mypar{Qualitative Results.}
To better understand how the assistants react to human input, Fig.~\ref{fig:examples_full_system_1} shows two examples of the collaborative annotation process in our full system. Notice the contextual response of the assistant: In the top row, once the annotator has annotated the small \emph{baseball glove}, our assistant changes many regions in the background to \emph{playingfield}.
In the bottom row, after the annotator changes the label of the ground from \emph{sand} to \emph{snow}, our assistant corrects the label of the {\em rocks} in the background and also adds a \emph{snowboard} as well as other snow regions. 

\mypar{Annotation quality in context.}
To get a better understanding of what 0.71 panoptic annotation quality means, we can put this into context by comparing to MCG~\cite{ponttuset17pami}, one of the best region proposal methods.
To enable this comparison, we re-evaluate quality in terms of Intersection-over-Union averaged over ground-truth regions (mIoU)~\cite{kirillov19cvpr,ponttuset17pami,uijlings13ijcv,xu16cvpr}.
Our 0.71 panoptic quality translates to 77.1\% mIoU. In contrast, MCG~\cite{ponttuset17pami} reports a 60\% mIoU on the same dataset, when selecting the best available proposal for each ground-truth region (out of 4000).
Moreover, we also compare the annotation quality of our system to manual polygon drawing through human experiments in Sec.~\ref{sec:human}.

\subsection{Study with human annotators}
\label{sec:human}

We now verify whether the efficiency gains observed in simulation also transfer to human annotators.

\mypar{Protocol.}
We randomly select 100 COCO validation images and have four humans annotate them (two experts and two non-experts).
We compare polygon drawing, fluid annotation~\cite{andriluka18acmmm}, and our system with all assistants.
Before starting the real experiment, all subjects practice for 30+ minutes on both the system with assistants and the system without. They also practice about 10 minutes with the polygon drawing interface (which is simpler).

Importantly, we want to avoid measuring confounding effects, such as annotators being inherently faster than others, or being faster because they already saw the image before.
We therefore make sure that each annotator annotates an equal number of images with each interface, while never annotating an image more than once.
Additionally, we want to avoid measuring human label disagreement as a confounding factor. The influence of such disagreement is serious, as the COCO group demonstrated by annotating 5000 COCO images twice by independent annotators~\cite{kirillov18coco}.
To rule out this factor, we show annotators the original COCO ground truth during annotation on one monitor, and ask them to reproduce it using the interface on another monitor.
Thanks to all the precautions above, our study can measure time differences due to the interface only.

\begin{figure}[t!]
\vspace{-.0cm}
  \includegraphics[width=\linewidth]{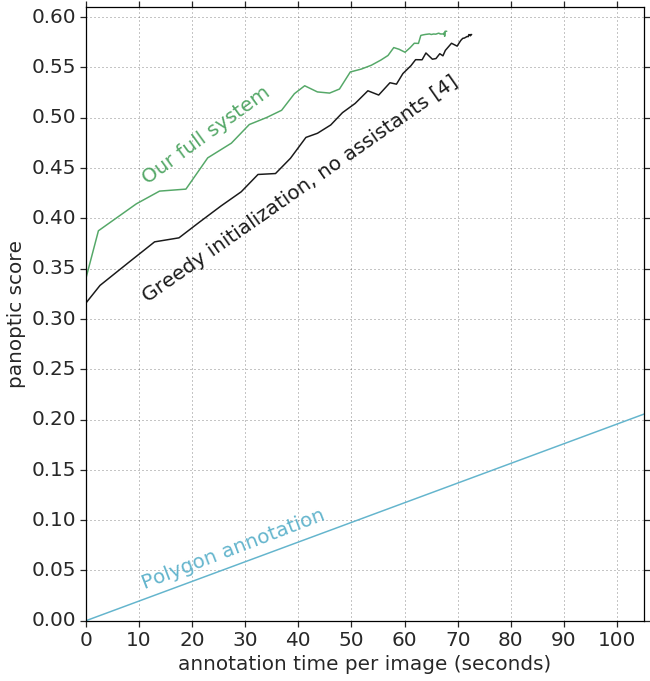}
  \vspace{-.3cm}
  \caption{Results of our user study on 100 images from the COCO panoptic dataset~\cite{caesar18cvpr,lin14eccv,kirillov19cvpr}.}
  \label{fig:human_results}
\end{figure}

\mypar{Results.}
Fig.~\ref{fig:human_results} shows quality (panoptic score) as a function of human annotation time (in seconds). 
Overall, our method emerges as more efficient than~\cite{andriluka18acmmm} and polygon drawing.
For example, after 40 seconds per image, the panoptic quality is 0.08 when doing polygon annotation, 0.48 for Fluid Annotation~\cite{andriluka18acmmm}, and 0.53 for our full system. Thus, in this low-time-budget regime, our approach makes a further improvement over~\cite{andriluka18acmmm}, which was already delivering much better annotations than polygon drawing.
These reported improvements match what we observed in the simulation experiments (Sec.~\ref{sec:eval_full}).

Polygon drawing requires 338s seconds on average to reach the end of the annotation process, compared to 67s for our method. Thus our method is $5\times$ faster. In terms of quality, it is important to note that even when the annotator draws polygons while looking at the original ground truth, they are unable to perfectly replicate the original annotation. In practice, they reach 0.74 panoptic quality, which can be seen as an upper-bound for this task. In comparison, when using our method, annotators reach 0.59 panoptic quality.

\mypar{Boundary Refinement Stage.}
So far, our system does not allow the annotator to refine the boundaries of the segments in the pool generated by Mask R-CNN.
But if we want to reach the highest annotation quality possible, we can add a post-processing refinement stage.
We did this by using a free-form drawing tool with adjustable brush size (similar to MS-paint) to refine the boundaries in the 100 images in this study.
Our resulting overall pipeline took a total of 139s per image to reach a panoptic annotation quality of 0.73 on average (67s for the main stage, plus 72s for the boundary refinement stage).
Hence, our overall pipeline is $2.4\times$ faster than manual polygon drawing while achieving nearly the same panoptic quality of 0.74.

\begin{figure}[t!]
\vspace{-.0cm}
  \includegraphics[width=\linewidth]{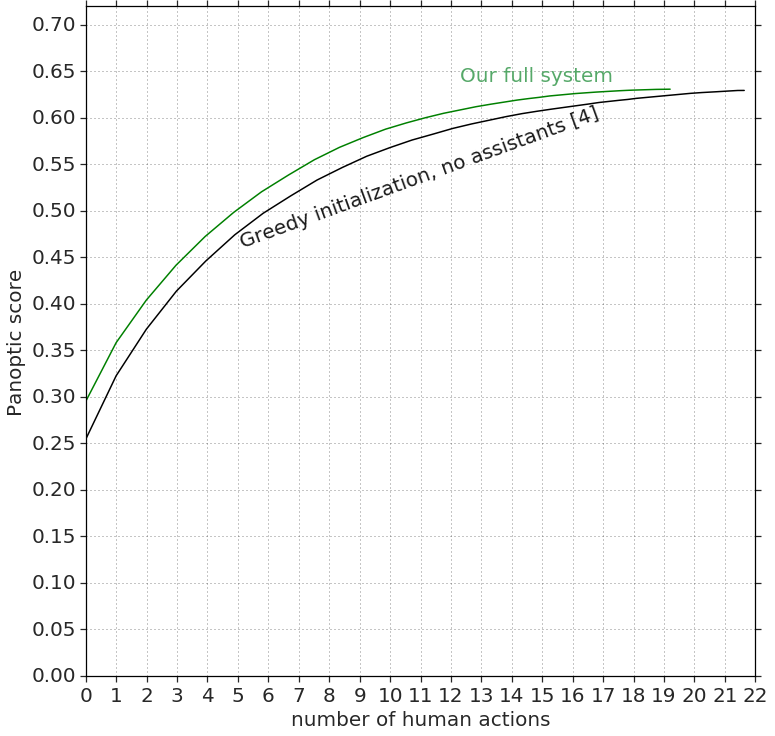}
  \caption{Simulated annotation results on the ADE20k validation set.}
  \label{fig:results_ade20k}
\end{figure}

\subsection{Annotating a new dataset}
\label{sec:cross_data}

So far we applied our system to annotate new images from the same dataset it was trained on. In practice though, often one wants to
annotate an entirely new dataset, with different appearance distribution and even new object and stuff classes. Here we explore this scenario by annotating ADE20k~\cite{zhou19ijcv} (in simulation).

We fine-tune our system on a very small amount (5\%, 1k images) of the ADE20k training set, so that it still brings a strong gain in overall annotation time. We use this data to train the underlying Mask-RCNN model for generating proposal sets and our assistants.
We fine-tune models starting from those pre-trained on COCO, and use a 4-fold leave-one-out strategy where the assistants are trained on different images than those used to train Mask R-CNN. This trains better assistants while making efficient use of the limited training data.

We evaluate our system on the validation set of ADE20k (2k images). As Fig.~\ref{fig:results_ade20k} shows, our results in Sec.~\ref{sec:eval_full} generalize to annotating ADE20k:
at 0.6 panoptic score, our system is 19\% faster than~\cite{andriluka18acmmm}.
Furthermore, our system reaches the final 0.63 panoptic quality 12\% faster than~\cite{andriluka18acmmm}.

\section{Conclusions}

We introduced a framework in which a human annotator and an automated assistant collaboratively annotate an image. The assistant intelligently reacts to human input based on context to annotate other parts of the image by itself. On the COCO panoptic dataset~\cite{caesar18cvpr,kirillov19cvpr,lin14eccv}:
(1) in simulation we demonstrated that our system is 17\%-26\% faster than the recent interface of~\cite{andriluka18acmmm};
(2) a human experiment confirms these improvements. At the same time, it shows that our method is $5\times$ faster than polygon drawing at a small loss of annotation quality. Furthermore, after adding a manual refinement stage, our system is $2.4\times$ faster without any compromise on quality.
In a cross-dataset experiment on ADE20k~\cite{zhou19ijcv}, (3) we show that starting from a very small amount of annotated data, our approach can be used to quickly annotate new datasets.

\bibliographystyle{ACM-Reference-Format}
\balance
\bibliography{shortstrings,loco,additional_refs}

\end{document}